\pdfoutput=1
\documentclass[11pt,a4paper]{article}
\usepackage[hyperref]{emnlp2020}
\usepackage{times}
\usepackage{latexsym}
\usepackage{amsmath}
\usepackage{array}
 \usepackage{graphicx}
\usepackage{multirow}
\usepackage{xcolor}
\usepackage{float}

\newcommand{\baseline}{2M-BASE~}
\newcommand{\threem}{3M~}
\newcommand{\twompgen}{2M-PGEN~}
\newcommand{\twompgennegation}{2M-PGEN-NEG~}
\newcommand{\threemnegation}{3M-NEG~}
\newcommand{\threempgennegationconcept}{3M-PGEN-NEG-CONCEPT~}

\usepackage{microtype}

\aclfinalcopy % Uncomment this line for the final submission
\def\aclpaperid{***} %  Enter the acl Paper ID here

\title{Dr. Summarize: Global Summarization of Medical Dialogue\\ by Exploiting Local Structures}

\author{Anirudh Joshi \\ Stanford University
\And
Namit Katariya \\Curai
\And
Xavier Amatriain \\Curai
\And
Anitha Kannan \\Curai
}

\date{}

\begin{document}
\maketitle
\begin{abstract}
Understanding a medical conversation between a patient and a physician poses unique natural language understanding challenge since it combines elements of standard open-ended conversation with very domain-specific elements that require expertise and medical knowledge. Summarization of medical conversations is a particularly important aspect of medical conversation understanding since it addresses a very real need in medical practice: capturing the most important aspects of a medical encounter so that they can be used for medical decision making and subsequent follow ups.

In this paper we present a novel approach to medical conversation summarization that leverages the unique and independent local structures created when gathering a patient’s medical history. Our approach is a variation of the pointer generator network where we introduce a penalty on the generator distribution, and we explicitly model negations. The model also captures important properties of medical conversations such as medical knowledge coming from standardized medical ontologies better than when those concepts are introduced explicitly. Through evaluation by doctors, we show that our approach is preferred on twice the number of summaries to the baseline pointer generator model and captures most or all of the information in 80\% of the conversations making it a realistic alternative to costly manual summarization by medical experts.

\end{abstract}

\definecolor{darkgreen}{rgb}{0.0, 0.42, 0.24}
\definecolor{amethyst}{rgb}{0.6, 0.4, 0.8}

\section{Introduction}
Telemedicine is a rapidly growing medium of interaction with the healthcare system \cite{telemedicine}. With the COVID-19 pandemic limiting in person medical visits, healthcare systems are seeing greater than 100\% increase in virtual urgent care visits and greater than 4000\% increase in 
\begin{figure}[H] %{\columnwidth}
\renewcommand{\arraystretch}{1.5}
\small
\begin{tabular}{ |p{7cm}| } 
\hline
\textbf{DR}: You mentioned \textbf{\textcolor{darkgreen}{having a cough for 2 days}} and a \textbf{\textcolor{darkgreen}{fever since last night}} along with \textbf{\textcolor{darkgreen}{being short of breath}}. Is that correct?

\textbf{PT}: \textbf{\textcolor{darkgreen}{yes}} , correct

\textbf{DR}:  I appreciate your concern for preventing spread. Do you feel like you are unable to move around as usual?

\textbf{PT}: I'm \textbf{\textcolor{darkgreen}{definitely weaker and low energy}} the \textbf{\textcolor{amethyst}{fever went down to 99 this morning}}

\textbf{DR}:  Have you taken any medications or tried anything else to help you with your symptoms?

\textbf{PT}: \textbf{\textcolor{amethyst}{lots of fluids and vitamin c. lozenges to minimize coughing}}

\textbf{DR}:  do you \textbf{\textcolor{red}{have any medical conditions}} or have you been on any medications

\textbf{PT}: \textbf{\textcolor{red}{no}}, none

\textbf{DR}: alright. When you had a fever, did you \textbf{\textcolor{red}{take medicine}} like \textbf{\textcolor{amethyst}{tylenol}} to bring the fever down? 
 
\textbf{PT}: I \textbf{\textcolor{red}{didn't}}	\\
\hline

\textbf{Model Output Summary}
\begin{itemize}
\item mentioned \textbf{\textcolor{darkgreen}{having a cough}} for 2 days and \textbf{\textcolor{darkgreen}{a fever}} since last night along with being \textbf{\textcolor{darkgreen}{short of breath}}.

\item unable to move around as usual. \textbf{\textcolor{darkgreen}{definitely weaker and low energy}} \textbf{\textcolor{amethyst}{fever went down to 99}}

\item \textbf{\textcolor{amethyst}{lots of fluids and vitamin c. lozenges}} to minimize coughing with symptoms .

\item \textbf{\textcolor{red}{no medical conditions.}} none have any medical conditions.

\item \textbf{\textcolor{red}{ didn't take medicine}} like \textbf{\textcolor{amethyst}{tylenol}} to bring the fever down.
\end{itemize} \\
\hline
\end{tabular}
\caption{Example medical dialogue and summary generated by our proposed model. Note that the summary captures \textbf{\textcolor{darkgreen}{affirmatives}}, \textbf{\textcolor{red}{negatives}} and \textbf{\textcolor{amethyst}{medical concepts}}. For more examples, see appendix.}
\label{fig:example}
\end{figure}
 virtual non-urgent care visits \cite{telemedicine}. Telemedicine systems today involve either direct voice and video chat or text based chat interfaces. At the end of a history taking conversation with the patient (\textit{i.e.} gathering of presenting symptoms, patient concerns and the past medical, psychological and social history), a doctor or nurse typically summarizes the information from the dialogue in order to pass it on to other care providers or as a means of recording the interaction. Even in more traditional in-person medical settings, the advent of electronic health records (EHR) as a way to record medical information has created the need to summarize any patient conversation with healthcare professionals. In all of the above cases, requiring human experts to summarize a potentially long medical conversation is costly and limits the scalability of the healthcare system. Furthermore, nurses or doctors who are required to do these tasks feel burnt out since the task feels repetitive and mechanical \cite{physicianburnout}.

Medical Dialogue summarization possesses unique characteristics and goals that are not present in other domains. Notes written by domain experts need to capture important parts of the conversation needed for clinical decision making, and not the summary of the entire conversation. As an example from Figure \ref{fig:example} we observe that from the conversation it is important to (1) capture all the medical conditions and terminology described in the dialogue (cough, fever, shortness of breath etc.) (2) discern all the affirmatives and negatives on medical conditions correctly (no allergies, having a cough for 2 days) and (3) bias towards copying from the source text while not being completely extractive. We observe that the majority of the information that is needed in a summary note is present in the medical dialogue with some novel words introduced to stitch phrases together. Unlike open domain dialogue between peers which involves long term memory dependencies between turns in the dialogue, patient history taking possesses an inherent local structure.

Another important challenge of end-to-end medical dialogue summarization is the lack of large scale annotated datasets. Annotations of medical dialogue needs trained doctors, which is expensive and slow. It is important to design modeling strategies and data capturing processes in a way that enables learning important biases as described above from sparse data.

In this paper, we propose a method to automate the generation of summary notes from the original patient/provider dialogue. Given the lack of existing public datasets with specific medical dialogues, we build our own dataset using conversations from a telemedicine platform and obtain reference summaries from healthcare professionals.

Our approach is based on the following key insights:
    \begin{enumerate}
        \item The learning problem for patient history taking summarization can be posed in the form of summarizing local dialogue turns (snippets) which are composed of smaller sections of the conversation. For example in Figure~\ref{fig:example}, the doctor's question about medical conditions and the patient's response can be considered a local snippet. \vspace{-.1in}
        \item Some specific characteristics of the medical conversation that are important for a doctor such as concept negations need to be explicitly modeled to avoid information loss.
    \end{enumerate}
    
Our proposed model leverages the pointer generator network \cite{pointergen} to capture the inductive bias present in our data. We extend the baseline pointer generator network by introducing a penalty to the generator distribution to guarantee that the network defaults to extractive summarization when necessary. We also model the unique domain specific challenges of medical data by introducing explicit modeling of negations and introducing medical concepts. Given our smaller dataset we found transformer based models produce poor outputs and were not pursued further. Recent work in medical summarization has also shown that Pointer Generator Networks produce strong performance \cite{radiology, krishna2020generating}. 

We propose a set of automated metrics that evaluate the dialogue on those particular challenges. We show through those automated metrics and doctor evaluations that the modified pointer generator network with the generator penalty is able to capture domain nuances and provide useful summarizations in over 80\% of the cases. While explicitly modeling negations introduces subtle improvements, those are not captured by our experts, and explicitly introducing medical concepts does not improve the performance of our model.

The main contributions of our paper can be summarized in:
\begin{enumerate}
    \item A novel end-to-end approach to medical dialogue summarization that is competitive with expensive and not scalable manual summaries.
    \item A simple extension of the pointer generator network that shows that adding a penalty for using the generator distribution has significant advantages in dialogues with high domain expertise like medicine
    \item A thorough evaluation of different summarization models that shows the correlation (and lack of such) between automated metrics and expert judgement

\end{enumerate}

%The pointer generator network has the right inductive biases we care about since it defaults to extractive summarization when necessary. 

\section{Related Work}
\noindent {\bf Neural Summarization}: Emergence of sequence to sequence models and attention mechanisms \cite{seq2seq,cnndm2} has led to rapid progress on extractive \cite{extract} , abstractive \cite{cnndm2} and hybrid models \cite{pointergen, copynet} for summarization. Much of the recent work has shown these models to generate near-human coherent summaries while retaining reasonable factual correctness. Of interest, is the class of hybrid models, that has inductive bias for being more extractive while possessing the ability to be abstractive for document text summarization tasks. Notably, \cite{pointless} introduced the idea of using dropout mechanism and pointer losses for this trade-off. In medical conversation summarization, harnessing the inductive bias of these hybrid models lead to more factually correct summaries, as we study in this paper. 

\noindent {\bf Dialog Summarization}: While most neural summarization has focused on news corpora, recent work has tried to tackle unique challenges associated with summarizing dialogues. \cite{gated} proposes using dialogue history encoders labelers based on type of dialogue section to inform the generation. \cite{didi} propose using key points as a means of categorizing sections of dialogue.

\noindent {\bf Medical Summarization}: \cite{extractive} explore the upper bounds on extractive summarization in medical text and find that a purely extractive approach may not provide sufficient recall. Incorporating medical knowledge into sequence to sequence summarization was studied by \cite{radiology} by encoding background information to condition the decoder. \cite{topicaware, soap} study spoken dialogue summarization in the medical domain with pointer generator networks however they don't explicitly model for the properties of medical data and do not report doctor evaluations of the outputs.

Our work differs by leveraging the unique local structures created when gathering a patient's medical history. We also explicitly incorporate into the learning process several properties of medical conversation that are important in summarization. While the pointer generator model shows a bias towards copying when trained on news corpora, we find that this is not true when trained on dialogue and our work on explicitly modulating generation probabilities to encourage copying has broad applicability to dialogue summarization in domains where factual correctness is important.

%Medical summarization tasks have their own challenges due to lack of sufficient amount of training data, while at the same time importance of correctness.

\section{Model}
We are interested in a model that has two main properties. First, it encourages copying from the snippet to preserve the integrity of the symptoms and medical issues being discussed. Second, it can handle out-of-vocabulary terms, such as medically relevant terms, that are used by patients and doctors.

The pointer generator \cite{pointergen} is naturally suited as they imbibe these properties by providing a hybrid of extractive and abstractive summarization,
with more emphasis on extraction \cite{pointless}.  We use pointer generator as the base model to build upon and encode medical conversation specific properties.

%This inductive bias described is desirable for the task of medical dialogue summarization. 

\subsection{Base model: Pointer Generator Network}
Pointer generator network \cite{pointergen} is a recurrent neural network based sequence model with attention and a  soft switch variable  $p_{\text{gen}}$ to 
orchestrate between copy and generation.  At each time step of decoding, the model uses  $p_{\text{gen}}$ to either copy words from the source text
using a pointer mechanism or generate words from a fixed vocabulary using the decoder probability distribution,
\begin{equation}
    P(w) = p_{\text{gen}} P_{\text{vocab}}(w) + (1-p_{\text{gen}}) \sum_{i:w_i=w}a^t_i \
\end{equation}
\begin{equation}
    {\text{loss}}_t= -\log{P(w^*_t)} + \lambda\sum_{i} \text{min}(a^t_i,c^t_i)
\end{equation}
 $P_{\text{vocab}}$ is a probability distribution over all words in the vocabulary.   $a^t$ is the attention/probability distribution over the source words that tells the decoder where to look to produce the next word. Thus, $P(w)$ the probability distribution over the extended vocabulary that is the union of fixed vocabulary and  the words that appear in the source, enabling the the model to copy out of vocabulary words. The loss in equation 2 encompasses the cross entropy and coverage loss described in \cite{pointergen}

\subsection{Incorporating medical knowledge}
\label{subsec:concept}
In the absence of sufficient amount of labeled data, transfer learning from a pre-trained model such as pointer generator may also be impoverished in distinguishing important ({\it  e.g} medical concepts) and unimportant out-of-vocabulary words. To infuse some of this knowledge,
we leverage compendium of medical concepts, known as unified medical language systems(UMLS).

During training, we use a one-hot vector $m^t$ that is same dimension as the source snippet. This vector encodes the presence of UMLS medical concepts that are in both source snippet and in reference. The requirement for presence of concepts in reference is to make sure that only those concepts that are relevant 
for the snippet is taken into account. $m^t$  influences the attention distribution  $a^t = \text{softmax}(e^t)$, through the functional form: 
\begin{equation*}
 e^t_i = v^t~\text{tanh}(W_hh_i + W_ss_t + w_cc^t_i + w_m{m}_i^t + b_{\text{attn}})
\end{equation*}
where $h_i$ is the encoder hidden state, $s_t$ is the decoder hidden state and $c_t$ is the coverage vector described in \cite{pointergen}. Since the concepts are encoded based on whether they are present in the reference, this acts as a form of teacher forcing where the concepts are encoded and supervised during training time but at test time, these encodings are not available to the model. 

Analogous to the coverage mechanism introduced by \cite{pointergen}, we propose to model this both, in the attention mechanism as well as in the loss function to directly supervise the model such that higher attention weights are placed on positions where concepts are present, by adding additional 
term  $\lambda_m(1 - \sum_{i}{m^t_i \cdot a^t_i})$ to the loss function described in equation 2. $\lambda_m$ is the scaling factor on the concept loss term and $a \cdot b$ is the dot product between $a$ and $b$.

%\begin{equation*}
%\begin{split}
%{\text{loss}}_t= &-\log{P(w^*_t)} + \lambda\sum_{i} \text{min}(a^t_i,c^t_i)~\TODO{b} \\ & + \lambda_m(1 - \sum_{i}{m^t_i \cdot a^t_i})
%\end{split}
%\end{equation*}
%where 

\subsection{Modeling negations}
\label{subsec:negation}
%Another critical component in medical dialogue summarization is to negate medical concepts and symptoms correctly, as described in the source dialogue. \TODO{refer to examples introduced in the intro -- if we sufficiently describe need
%for negation modeling in intro, we can skip `why' here} While this could be learnt implicitly given enough training data, we propose to model this %explicitly for a low-data scenario.

We take two complementary approaches to `supervise' modeling of negations - attention mechanism on negation words, and by explicitly modeling a switching variable that induces a mixture model over copy, generate and negate. 

\noindent {\bf Negation word attention:}  Similar to modeling of medical concept, negation attention directly supervises the model in the attention distribution and in the loss function. For this, a small set of negative unigrams ( `no' , `nope', `doesn't', `not' )  are manually curated. An additional binary vector ${\bf n}^t$ of the same length as that of the source snippet encodes $n_i^t=1$ when $t^{th}$ location in the source has one of these negative unigrams. The attention distribution is modified to focus the attention distribution on such terms.
\begin{equation*}
\begin{split}
e^t_i ={} v^t~\text{tanh}(&W_hh_i + W_ss_t + w_cc^t_i +{} \\ &w_m{m}_i^t + w_n{n}_i^t + b_{\text{attn}})
\end{split}
\end{equation*}
The loss function is augmented with $\lambda_n (1 - \sum_{i}{n^t_i \cdot a^t_i}$

%Similar to concept attention loss, we use an additional term called the negation attention loss \[k(1 - \sum_{i}{n_{\text{attn}}}^t_i)\] where ${n_{\text{attn}}}$ is the dot product between the negation encoding and the attention mechanism to promote higher weights on negation words.

%(Need better way to phrase this )Based on the experimental evidence, it was clear that a more aggressive method of modeling negations was needed.
\noindent {\bf Negation as a switching variable:}
%\TODO{Explain rationale for this, in addition to negation attention} 
In addition to the snippet-level summary, we also collect explicit labels in the form of a special token `[NO]' for parts of the snippet that are negated. While the [NO] token can be added to the fixed vocabulary, we note that the model would need to learn when to generate the [NO] token in the final summary using the decoder, and thereby influencing the likelihood of $p_{\text{gen}}$ in other abstractive parts of the summary. Instead, we use this additional signal to formulate the probability distribution over extended vocabulary as a convex combination:
\begin{equation*}
    P(w) = p_{\text{gen}}P_{\text{vocab}}(w) + p_{\text{copy}}\sum_{i:w_i=w}a^t_i +  p_{\text{neg}}P_{\text{[NO]}}
\end{equation*}
where $p_{\text{neg}}$ controls the generation of [NO]. This extends \cite{pointergen} with additional switching variable $p_{\text{neg}}$:
\vspace{-.1in}
\begin{equation*}
p_{\text{gen}},~p_{\text{copy}},~p_{\text{neg}} =\text{softmax}(w^T_{h^*}h^*_t + w^T_{s}s_t +{}  w^T_{x}x_t + b_{ptr})
\end{equation*}
where $h^*_t$ is the context vector, $s_t$ is the decoder state and $x_t$ is the decoder input. In positions where $p_{\text{neg}}$ is 1, $p_{gen} + p_{copy}$ needs be 0, and vice versa so that [NO] token is correctly incorporated into the summary during decoding.  We explicitly supervise this behavior by adding an additional L1 loss term to encourage $|p_{neg} -  (p_{gen} + p_{copy})|$ to be maximal. The L1 loss is weighted by a scalar factor $\gamma$ to modulate its contribution.

%Since these models utilize 3 mixtures to model the final probability distribution we refer to them as 3-mixture models or 3M.

\subsection{Controlling generation probability}
\label{subsec:pgen}

%\TODO{\cite{pointergen} observe that while the pointer generator network ends training with $p_{gen}$ at 0.53, $p_{gen}$ is far lower (0.17) at inference. }

% of news articles usually needs ``sequential'' copying i.e. parts of the article that need to be copied appear sequentially without having to be revisited. Therefore, this is what a pointer generator network \cite{pointergen} trained on news corpora such as CNN-Daily Mail \cite{cnndm1, cnndm2} would learn. While perfectly valid for news articles, figure~\ref{fig:pgen} illustrates how medical dialogue summarization requires the model to go back and forth between the doctor questions and patient answers to copy the complete summary.

%While the pointer generator network can handle this by switching between generation and copying, 
%This is hard for a pointer generator network to learn, especially when there is only a small amount of data.  Rather than learning to constantly shift the copy distribution the model learns to use the generator more.

We explicitly want summaries to be copied as much as possible from the source dialogue since factual errors in a medical setting are unacceptable. In order to do this copying, we need copy distributions that can shift rapidly back and forth between doctor and patient turns, while still {\it generating} to stitch between them. See table.~\ref{fig:pgen} for an example.

As the pointer generator has flexibility to switch between generation and copying, it has the (dis)advantage of using  $p_{\text{gen}}$ to compensate for lack of flexibility to rapidly shift between copy and generated by mostly depending on generation. In fact, we found this behavior to be true empirically: In \cite{pointergen}, at inference time, for text summarization task, $p_{\text{gen}}$ is below 0.2. In zero-shot setup, average $p_{\text{gen}}$ on our medical dialogue dataset is 0.2. However after fine-tuning on our dataset, we observe that the average $p_{\text{gen}}$ is 0.4 at inference time validating that the model depends a lot more on generation for conversation summarization.

This is detrimental since the model can choose to hallucinate medical concepts that are not part of the snippet.  We propose a penalty on the model for using the generator distribution instead of the copy distribution to force it to learn to use the copy mechanism effectively. We add $\delta~ p_{\text{gen}}$ term to the overall loss where $\delta$ is a scalar constant. This term will be large if the model is using the generator more during decoding.

\section{Evaluation}
We evaluate models using automated metrics and manual evaluation from doctors. Multiple studies have shown that automated metrics in NLP do not always correlate well to human judgments as they may not fully capture coherent sentence structure and semantics \cite{fair, salesforcemetrics}. Since medical dialogue summarization would be used to assist health care, it is important for doctors to evaluate the quality of the output.

\subsection{Automated metrics}
While we measure model performance on standard metrics of ROUGE \cite{lin-2004-rouge}, we also wanted to specifically measure a model's effectiveness in capturing the  medical concepts that are of importance, and the negations. Therefore, we propose a new set of automated metrics that directly measure medically relevant information in the summary.

\noindent {\bf Medical Concept Coverage}: The concept coverage set of metrics captures the encapsulation of the  medical terms in the model's output summary to the ground truth reference. In particular, let $\mathcal{C}$  be the set of medical concepts in the reference summary and $\hat{\mathcal{C}} $ be the set of concepts in the summary output by the model. Then,
$\textrm{Concept recall} = \frac{\sum_{n=1}^{N}  |\hat{\mathcal{C}}^{(n)}  \cap  \mathcal{C}^{(n)}| }  {\sum_{n=1}^{N}|\mathcal{C}^{(n)}|} $ and $\textrm{Concept precision} = \frac{\sum_{n=1}^{N}  |\hat{\mathcal{C}}^{(n)}  \cap  \mathcal{C}^{(n)}| }  {\sum_{n=1}^{N}|\hat{\mathcal{C}}^{(n)}|}$

%\begin{eqnarray}
 %   \textrm{Concept recall} &=& \frac{\sum_{n=1}^{N}  |\hat{\mathcal{C}}^{(n)}  \cap  \mathcal{C}^{(n)}| }  {\sum_{n=1}^{N}|\mathcal{C}^{(n)}|} \\
  %  \textrm{Concept precision} &=& \frac{\sum_{n=1}^{N}  |\hat{\mathcal{C}}^{(n)}  \cap  \mathcal{C}^{(n)}| }  %{\sum_{n=1}^{N}|\hat{\mathcal{C}}^{(n)}|}
%\end{eqnarray}

We use these to compute a Concept F1. We use an inhouse medical entity extractor to match concepts in the summary to UMLS. Medical concepts in the decoded summary that weren't present in the original conversation would be false positives and vice versa for false negatives.

\noindent {\bf Negation Correctness}:  To measure the effectiveness of the model to identify the negated status of medical concepts,
we use Negex \cite{negex09} to determine negated concepts. Of the concepts present in the decoded summary, we evaluate precision and recall on whether the decoded negations were accurate for the decoded concepts and compute a Negation F1. 

\subsection{Doctor Evaluation}
\label{sec:doceval}
We also had two doctors, who serve patients on our telehealth platform, evaluate the summaries produced by the models. Given the local dialogue snippets and the generated summary, we asked them to evaluate the extent to which the summary captured factually correct and medically relevant information from the snippet. Depending on what percentage of the concepts were correctly mentioned in the decoded summary of the provided snippet, the doctors graded the summaries with \emph{All} (100\%), \emph{Most} (at least 75\%), \emph{Some} (at least 1 fact but less than 75\%), \emph{None} (0\%) labels. We also formulated a comparison task where given two summaries generated by different models and the associated dialogue, they were asked which summary was better. The doctors also had the ability to use ``both'' and ``none'' depending on if both models captured a good summary or if none of them did. To avoid bias, the doctors do not know the model that produced the summary in both the experiments. In the comparison task, the two summaries were provided in randomized order so that there is no bias in the order of presentation of the summaries.
\section{Dataset construction}
\label{sec:dataset}
We collected a random subset of dialogue of 25,000 conversations from a telemedicine platform.  We split the dialogue into a series of local dialogue snippets using a simple heuristic: the turns between two subsequent question by the physician corresponds to a snippet. The length of these snippets ranged anywhere from two turns (a physician question and patient response) to ten turns.

We had medical doctors summarize a random sample of 3000 snippets. These are the same doctors who practice on the same telemedicine platform. The doctors were asked to summarize the sections as they would for a typical clinical note by including all the relevant information. Further if the summary included a negated medical term, eg. ``doesn't have fever'', the doctors were asked to use a [NO] token in front of that particular sentence in the summary. If a local snippet did not contain any relevant information they were excluded from annotations. For example in the beginning or end of conversations there may be turns that are purely greetings and not part of the patient history taking process, or purely educational in nature.

At the end of the labeling, we used the 1690 number of local snippets that the doctors labeled as containing pertinent information for history gathering.
We used 1365 as the training set, 158 as a validation set and 167 as a held out test set. The test set was made sure to be from distinct conversations that were taken from a different date range on the platform compared to the training or validation sets. This was done to ensure that different local snippets from a certain conversation weren't part of the training and test sets. The data was preprocessed by removing the names of the actors in the dialogue (``Doctor'', ``Patient'') and concatenating all the turns within a snippet together.

%Since obtaining a larger labeled dataset is expensive and impractical we use the CNN-Daily mail corpus as a pretraining corpus prior to finetuning on our data.

\section{Experiments}

\noindent {\bf Model variants}: We study the following variants: %All of them are pretrained on CNN-DailyMail summarization task.
\begin{itemize}
    \item \textbf{\baseline}: Pretrained pointer generator model fine-tuned on medical dialogue summarization \vspace{-.1in}
    \item \textbf{\twompgen}: \baseline + generator loss to control generation probability (\S~\ref{subsec:pgen}) \vspace{-.1in}
    \item \textbf{\twompgennegation}: \twompgen + negation attention mechanism and loss (\S~\ref{subsec:negation}) \vspace{-.1in}
    \item \textbf{\threem}: Pretrained pointer generator model fine-tuned on medical dialogue summarization using negation as a switching variable to form a 3 mixture final probability distribution ( \S~\ref{subsec:negation})\vspace{-.1in}
    \item \textbf{\threemnegation}: \threem + negation attention loss (\S~\ref{subsec:negation})\vspace{-.1in}
    \item \textbf{\threempgennegationconcept}: \threemnegation + the losses (\S~\ref{subsec:pgen}) to control generation probability and \S~\ref{subsec:concept} to improve medical concept coverage.
\end{itemize}

\noindent\textbf{Training details}:
All the models use a vocabulary size of 50k with 128 dimensional embeddings and 256 dimensional hidden states. The training parameters followed \cite{pointergen} with a learning rate of 0.15 and Adagrad as the optimizer. The coverage mechanism as described in \cite{pointergen} was used for all our models. Models were first pretrained on the CNN-Daily Mail corpus and finetuned on conversational data from our in-house chat-based telehealth platform (\S ~\ref{sec:dataset}). 

Pretraining took approximately 2 days on a single NVIDIA Titan Xp GPU and finetuning took under 2 hours. The concept and negation attention modifications added 512 parameters each to the base pointer generator model with coverage. For details on hyperparameters and validation results see appendix.

\subsection{Main Results}

\begin{table*}[ht!]
\resizebox{\textwidth}{!}{
\begin{tabular}{m{12.5em}|m{1.5cm}|m{1.5cm}|m{1.5cm}|m{1.2cm}|m{1.2cm}|m{1.1cm}|m{0.9cm}}
\hline
\multirow{2}{*}{\textbf{Models}} & \multicolumn{3}{c|}{\textbf{Metrics}} & \multicolumn{4}{c}{\textbf{Doctor Evaluation}}\\
\cline{2-8}
 & \textbf{Negation F1} & \textbf{Concept F1} & \textbf{ROUGE-L F1} & \textbf{Model} & \textbf{Baseline} & \textbf{Both} & \textbf{None}  \\
\hline
\baseline & $70.1  \pm  0.8$ & $69.1  \pm  1.3$ & $52.6 \pm 0.9$ &- & - & -   & - \\
\hline
\twompgen & $67.3 \pm 3.3$ & $72.8 \pm 0.8$ & $55.4 \pm 0.9$ & 37.1\% & 18.5 \%  & 38.9 \% & 5.3\%\\
\hline
\twompgennegation & $72.2 \pm 3.6$  & $70.9 \pm 2.2$ & $53.5 \pm 0.7$ & 37.7\% & 22.7\% & 34.1\% & 5.4\%\\
\hline
\threempgennegationconcept & $78.0 \pm 4.2$ & $70.6 \pm 1.4$ & $55.2 \pm 1.2$  & 26.9\% & 25.7\%  & 42.5\% & 4.2\%  \\
\end{tabular}
}
\caption{Automated and Doctor evaluation.}
\label{tab:learnedlr}
\end{table*}

%We evaluated the different models considering the following three properties that we identified to be important in medical dialogue summarization.

%\TODO{this needs to align with metrics}

%1. Bias towards copying text from the snippet.

%2. Covering all medical concepts from the snippet.

%3. Preserving the correct negation / affirmation of medical conditions i.e. if a user has / doesn't have a symptom X, the summary should correctly reflect that. 

Table~\ref{tab:learnedlr} presents the key results, with a side-by-side comparison between automated metrics and doctor evaluation. We chose \twompgen as the default improvement over \baseline as it is the model with the simplest improvement over \baseline. The subsequent models build on \twompgen by modeling negations and explicit concept attention. We make the following observations:

% \begin{figure*}
%     \centering
%     \includegraphics[height=7cm,width=13cm]{sections/figures/examples.png}
%     \caption{Snippets along with multiple model outputs}
% \label{fig:examples}
% \end{figure*}

\begin{table}[]
\renewcommand{\arraystretch}{1.5}
\tiny
\begin{tabular}{p{0.15\textwidth}p{0.30\textwidth}}
\hline
\multicolumn{1}{c}{\textbf{Snippet}}                                                                                          & \multicolumn{1}{c}{\textbf{Model Summaries}}          \\
\hline
\multirow{4}{*}{
\begin{tabular}[c]{@{}p{0.15\textwidth}@{}}
\textbf{DR}: what color is the phlegm? \\ \textbf{PT}: dark green dark green
\end{tabular}} & \textbf{\baseline}: the color is the phlegm. \\ \cline{2-2} & \textbf{\twompgen}: the phlegm is dark green dark green \\ \cline{2-2} & \textbf{\twompgennegation}: phlegm color is dark green dark green. \\ \cline{2-2} & \textbf{\threempgennegationconcept}: has dark green dark green color. \\
\hline
\multirow{4}{*}{
\begin{tabular}[c]{@{}p{0.15\textwidth}@{}}
\textbf{DR}: thanks for letting me know. does your back hurt? \\ \textbf{PT}: nope my back feels fine
\end{tabular}} & \textbf{\baseline}: back pain does not back feels fine. \\ \cline{2-2} & \textbf{\twompgen}: back is back . thanks for letting me. \\ \cline{2-2} & \textbf{\twompgennegation}: no back hurt . fine back feels fine. \\ \cline{2-2} & \textbf{\threempgennegationconcept} [NO] no back hurt . back feels fine. \\
\hline
\multirow{4}{*}{
\begin{tabular}[c]{@{}p{0.15\textwidth}@{}}
\textbf{DR}: What was your last period date? \\ \textbf{PT}: feb 9th to 13th i have an average cycle of about 32-33 days.
\end{tabular}} & \textbf{\baseline}: last period is feb 9th to 13th. has an average cycle of 32-33 days. \\ \cline{2-2} & \textbf{\twompgen}: period date feb 9th to 13th. average cycle of about 32-33 days \\ \cline{2-2} & \textbf{\twompgennegation}: period feb 9th to 13th. has an average cycle of about 32-33 days \\ \cline{2-2} & \textbf{\threempgennegationconcept}: has an average cycle of 32-33 about feb 9th to 13th. has an average cycle of 32-33 days. \\
\hline
\end{tabular}
\caption{Snippets along with multiple model outputs}
\label{tab:examples}
\end{table}
\begin{itemize}
\item \twompgen improves concept F1 score over \baseline at the cost of drop in negation score. Even when we consider snippets where \baseline and \twompgen have identified the same set of concepts, we find that \twompgen generates better summaries. This is also evidenced by the corresponding doctor evaluation, where we see \twompgen preferred on twice the number of examples compared to the \baseline (37.1\% vs 18.5\%). 

Qualitatively, consider the first and third examples in Table~\ref{tab:examples} in which both models have identified the medical concepts like ``phelgm" and ``cycle" however \twompgen clearly provides more descriptive and coherent summaries explaining the wide margin on human evaluation. \vspace{-.1in}
\item With \twompgennegation model that extends \twompgen with negation attention, we can see that the negation F1 improves, with a dip in concept attention. While in doctor evaluation, both \twompgennegation and \twompgen perform comparably, in comparison to \baseline, we do find that on difficult negations such as the second example in Table~\ref{tab:examples}, models that explicitly incorporate negations prove better.\vspace{-.1in}
\item  \threempgennegationconcept performs best on the negation metric while maintaining comparable performance on concept metric to \baseline. However,  it is not our best performing model on human evaluation. We analyze this in detail in \S~\ref{sec:exp_medicalconcepts}. From the first and third examples in Table~\ref{tab:examples} we can see that sentence structure and coherency reduces on \threempgennegationconcept despite capturing concepts and negations.\vspace{-.1in}
\item Across all models, we find that there is only 5\% of the snippets where no model produces good summary. On closer investigation, we find that these are snippets where there is a lack of coherent response from the patient.
\end{itemize}

\subsection{Independent model evaluation by doctors}
\begin{figure}
    \centering
    \includegraphics[scale=.25]{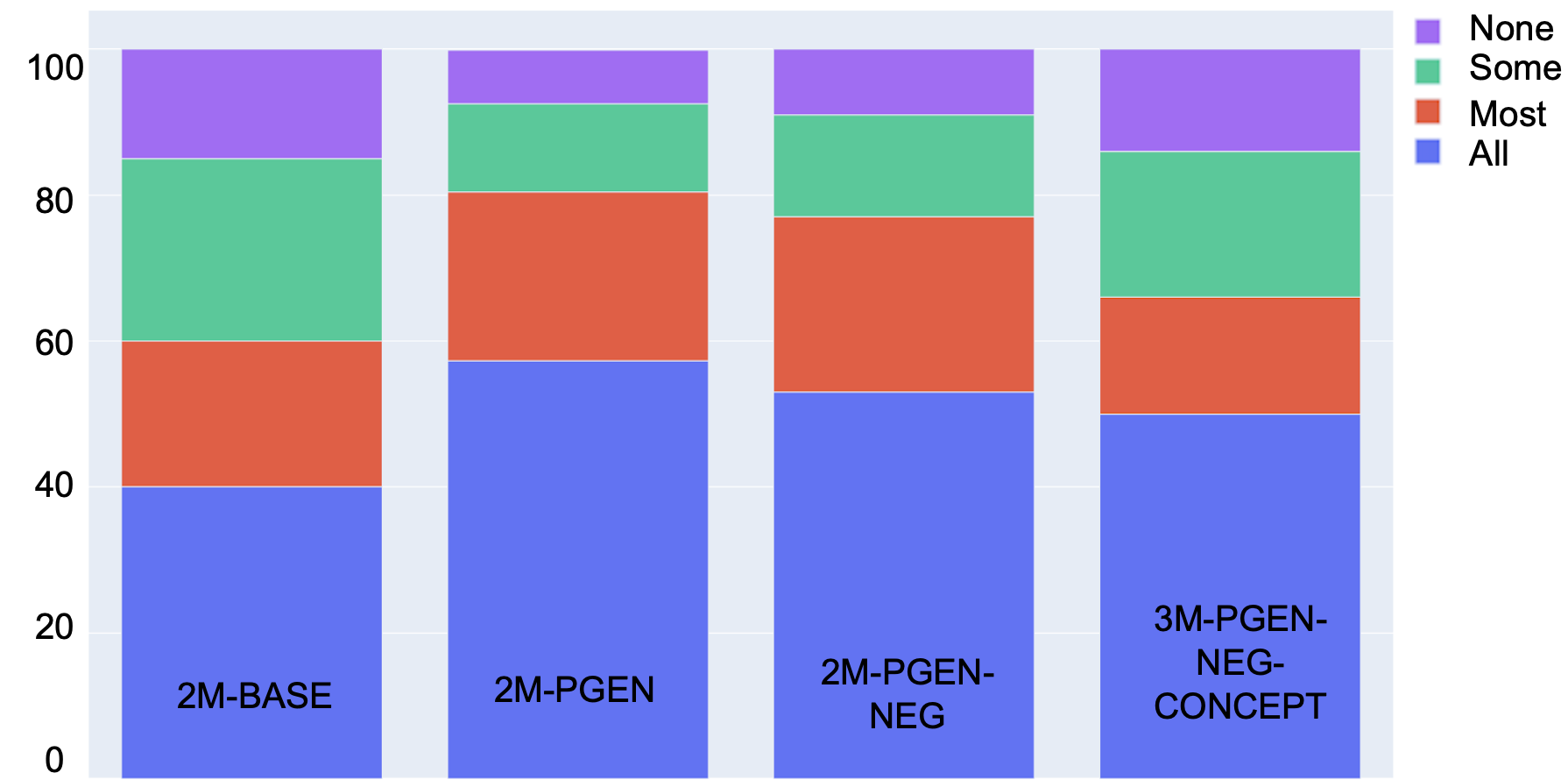}
    \caption{Doctor evaluation of amount of relevant information covered by summaries }
\label{fig:task1}
\end{figure}
%[height=5cm,width=8cm]
In Figure \ref{fig:task1}, we study doctor's evaluations when models were evaluated independently. The model's output summary was graded on whether it included ``All'', ``Most'', ``Some'', ``None'' of the relevant facts, \twompgen gets all or most of the facts on 80\% of examples compared to 60\% for \baseline. From Table~\ref{tab:examples} and Table \ref{fig:pgen} this is evident with the integrity of the patient response being preserved when using a generator penalty.

\begin{table*}[]

\centering
\resizebox{\textwidth}{!}{\begin{tabular}{lll}
%\toprule
\hline

\textbf{Snippet}   & \textbf{\baseline} & \textbf{\twompgen} \\ 
\hline
\begin{tabular}[c]{@{}l@{}}
DOCTOR : Thanks for letting me know. \\
  \hspace{.7in}How severe is the pain on a scale of 1-10, \\
  \hspace{.7in}considering 10 being the worst and severe? \\
PATIENT: 4-5
\end{tabular}  &
\begin{tabular}[c]{@{}l@{}}
\colorbox{pink}{pain} \colorbox{orange}{is} \colorbox{pink} {severe} \colorbox{orange}{.} 
\colorbox{pink}{pain} \colorbox{orange}{is} \\ \colorbox{pink}{10} \colorbox{orange}{on} \colorbox{pink}{a scale} \colorbox{orange} {of 1-10.}
\end{tabular}  &
\begin{tabular}[c]{@{}l@{}}
\colorbox{pink}{the pain is} \colorbox{green}{4-5} \colorbox{pink} {on a scale} \\ \colorbox{pink}{of 1} \colorbox{orange}{-} \colorbox{pink}{10 being the worst and severe}
\end{tabular}
\\ \hline

\begin{tabular}[c]{@{}l@{}}
 DOCTOR: Thanks for sharing. \\
  \hspace{.7in} Have you been on any medications for the same? \\
PATIENT: I’ve been on isotretenoin for the acne but off of it \\
  \hspace{.7in}for the last 5 months nothing for the bradycardia\\
\end{tabular}  &
\begin{tabular}[c]{@{}l@{}}
\colorbox{orange}{not} \colorbox{green}{on isotretenoin for acne} \colorbox{orange}{but}  \\ \colorbox{green}{off of bradycardia for last 5 months} \colorbox{orange}{.}
\end{tabular}  &
\begin{tabular}[c]{@{}l@{}}
\colorbox{green}{been on isotretenoin for the acne} \\ \colorbox{green}{but off of it for the last 5 months}\\

\end{tabular}
\\ \hline

\begin{tabular}[c]{@{}l@{}}
DOCTOR: does your chest pain get worse \\
  \hspace{.7in}while taking deep breaths ? \\
PATIENT: no 
\end{tabular}  &
\begin{tabular}[c]{@{}l@{}}
\colorbox{orange}{no} \colorbox{pink}{chest pain} \colorbox{orange}{. worse} \\ \colorbox{pink}{while} \colorbox{pink}{taking breaths} 
\end{tabular}  &
\begin{tabular}[c]{@{}l@{}}
\colorbox{green}{no} \colorbox{pink}{chest pain} \colorbox{orange}{get} \colorbox{pink}{worse} \\ \colorbox{pink}{while taking deep breaths}

\end{tabular}
\\ \hline

\end{tabular}
}

\caption{Modulating $p_{gen}$: Directly modulating the $p_{gen}$ allows for the copy mechanism to learn how to copy between doctor questions and patient answers instead of just sequentially copying. \colorbox{pink}{Pink} is used to show words copied from the Doctor, \colorbox{green}{green} for words copied from the Patient and \colorbox{orange}{orange} for words generated.}
\label{fig:pgen}
\end{table*}

\subsection{Role of explicitly modulating $p_{\text{gen}}$}
Table~\ref{fig:pgen} provides a qualitative comparison between \baseline  and  \twompgen. We observe that \baseline relies more on the generator to create the summaries and at times does not copy from the patient's answers. For instance, in the second example in  Table~\ref{fig:pgen}, `no' is produced by the generator in \baseline.  In contrast, \twompgen is able to  learn a  copy mechanism that shift attention distributions over to patient answers and back to doctor questions as opposed to linear copying. For the same example, `no' is copied from the patient.

We can also see examples in Table~\ref{fig:pgen} where \baseline is more erroneous than \twompgen. Additionally, while \baseline correctly identified all three concepts (acne, bradycardia and isotretenoin), \twompgen captured the semantics of summarizing that snippet, even though the resultant summary had only two of the three concepts (acne, isotretenoin). This example also sheds light on why automated metrics are not as reliable in measuring the efficacy of the models.

%As discussed in \S~\ref{subsec:pgen}, without explicit signal, \baseline learns the easier mechanism of using the copy mechanism sparingly while relying on the generator to build the summary. As seen in second example,  "no" is copied from the patient \twompgen while  In contrast, since $p_{\text{gen}}$ is penalized in \twompgen,  it is able to

%One reason why explicitly modulating $p_{\text{gen}}$ produces strong human evaluation outcomes is that summarizing dialogue involves guiding the copy mechanism to shift attention distributions over to patient answers and back to doctor questions as opposed to linear copying.  By penalizing $p_{\text{gen}}$ the model has to make the trade-off between using the generator and improving cross entropy loss. It can only reduce the overall loss using the generator distribution if the reduction in cross entropy loss compensates for the $p_{\text{gen}}$ penalty.

%From Table~\ref{fig:pgen} we observe that \baseline relies more on the generator to create the summaries and at times doesn't copy from the patient's answers. In the second example,
\subsection{Role of explicit negation modeling} 
%From Table~\ref{tab:learnedlr}, we observe that \twompgennegation improves Negation F1 over \twompgen (72.2 vs 67.3) and introducing \threem helps boost Negation F1 further. 

To determine which mode of explicit negation modeling has greater effect we compare \threem and \threemnegation against \baseline (Table \ref{tab:quant3m}). We observe that extending \baseline to \threem with the $p_{neg}$ soft switch improves Negation F1 (76.9 vs 70.1). Further extending this to \threemnegation by incorporating negation attention we see an even larger improvement in Negation F1 (81.5). We also find that on difficult negations such as the second example in Table~\ref{tab:examples} where the patient responds ``nope'' followed by an affirmative ``fine'', models that explicitly incorporate negations produce better summaries. However from the human evaluation and qualitative examples (Table 2) we see that coherency reduces even though the quantitative metrics improve. See appendix for qualitative comparison.
 %when using the negation attention and associated loss term.

% Since $p_{\text{neg}}$ controls the generation of [NO] and [NO] is sparsely prevalent in the reference, the ideal distribution of $p_{\text{neg}}$ over the length of decoding is to hit 1.0 during [NO] generation and 0 elsewhere. Supervising $p_{\text{neg}}$ explicitly with an L1 loss to maximise the margin between $p_{\text{gen}}$ + $p_{\text{copy}}$ and $p_{\text{neg}}$ helps enforce the model to learn this distribution of $p_{\text{neg}}$.

\begin{table}[ht!]
\begin{tabular}{m{2cm}|m{1.5cm}|m{2cm}}
\hline
\textbf{Model} & \textbf{Negation F1} & \textbf{ROUGE-L F1}   \\
\hline
\baseline & $70.1  \pm  0.8$
& $52.6 \pm 0.9$  \\
\hline
\threem   & $76.9 \pm 3.6$ & $56.4 \pm 0.3$ \\
\hline
\threemnegation & $81.5 \pm 4.7$ & $54.5 \pm 1.3$ \\
\hline

\end{tabular}
\caption{Negation ablation} %:  switching variable and word attention}
\label{tab:quant3m}
\end{table}
%\vspace{-.1in}
\subsection{Role of encoding medical concepts}
\label{sec:exp_medicalconcepts}
Given data sparsity our hypothesis was that directly using medical concepts to guide the attention mechanism would help performance on the concept metric. We see improvement in this metric when adding concept attention to \baseline (concept F1 72.0 vs 69.1) however once $p_{gen}$ loss is introduced we notice these gains no longer hold. On local snippets, we observe that the increased copying ability compensates for the removal of concept attention and adding concept attention can reduce performance. In both cases the attention on medical concepts in the copy distribution increases 5\% however this doesn't amount to consistent increase on the automated metrics. We leave this as an open research direction for longer dialogue snippets where enhanced copying may need to be coupled with concept attention.

%\TODO{are there stats we can provide on longer snippets here?}

\section{Conclusions}

In this paper, we presented a novel approach to medical conversation summarization. This is an important application for text summarization since medical professionals rely on good conversation summarizations for medical decision making and follow up. Medical conversations, however, have traditionally posed a challenge for vanilla machine learning approaches because of the importance of domain knowledge and syntactic nuances such as negation. We extend a deep learning approach, pointer generator networks, and show that for domains like medicine where integrity of the source is critical, encouraging copying in the learning process produces the best model (2M-PGEN) on human evaluation. This approach represents a viable alternative to human summarization since experts report that up to 80\% of the relevant information is present in 2M-PGEN summaries with only 5\% of summaries containing no relevant information. Even if the system implementing this approach could not operate completely automated, it is clear that it could speed up the summarization process by reducing the amount of human intervention needed.

For future work, we would like to see if our findings generalize well on other datasets and other domains. Particularly, we would like to see if the anecdotal evidence that explicit negation modeling matters, can be captured by the metrics or the expert human evaluation.

\bibliography{emnlp2020}
\bibliographystyle{acl_natbib}

\section{Appendix}
\label{sec:supplement}

\subsection{Hyperparameters}
The key hyperparameters introduced are the weighting factors used on the additional loss terms
%The hyperparameters were chosen to keep the loss terms within similar orders of magnitude.
$\lambda_m$, $\lambda_n$ and $\delta$ which were used to weight the concept, negation and $p_{\text{gen}}$ loss were set at 1.0, 0.1 and 1.0 to scale the contribution of those terms to be roughly within the same range. For experiments with \threem models, $\gamma$ the weight on the L1 loss term used to supervise $p_{\text{neg}}$ was set at 2.0. While the $p_{\text{neg}}$ supervision reduced the contributions of the other loss terms we found scaling up the other weights didn't have a material impact on the quantitative and qualitative performance. For simplicity we retained the same hyperparameters for all the models and runs. The bounds for each hyperparameter can be varied within an order of magnitude without significantly impacting the qualitative or quantitative metrics.

\begin{table*}
%\renewcommand{\arraystretch}{1.5}
%\tiny
\resizebox{\textwidth}{!}{
\begin{tabular}{m{12.5em}m{1.5cm}m{1.5cm}m{1.5cm}}
\hline
\multirow{2}{*}{\textbf{Models}} & \multicolumn{3}{c}{\textbf{Metrics}}\\
\cline{2-4}
 & \textbf{Negation F1} & \textbf{Concept F1} & \textbf{ROUGE-L F1}   \\
\hline
\baseline & $71.2  \pm  1.4$ & $70.2  \pm  1.1$ & $53.2 \pm 1.4$  \\
\hline
\twompgen & $76.9 \pm 0.5$ & $72.2 \pm 0.9$ & $55.4 \pm 0.6$ \\
\hline
\twompgennegation & $74.4 \pm 2.6$  & $71.3 \pm 1.5$ & $54.4 \pm 1.0$\\
\hline
\threempgennegationconcept & $77.3 \pm 2.1$ & $69.3 \pm 2.4$ & $53.8 \pm 2.9$   \\
\end{tabular}
}
\caption{Validation Performance}
\label{tab:validationmetrics}
\end{table*}

\begin{figure*}[!]
\renewcommand{\arraystretch}{1.5}
\begin{tabular}{ |m{\linewidth}| } 
\hline
\textbf{DR}: what was your last period date ?

\textbf{PT}: feb 9th to 13th I have an average cycle of about 32-33 days.

\textbf{DR}: are/were you on any hormonal form of birth control apart from plan b?

\textbf{PT}: no

\textbf{DR}:  Thanks for letting me know.

\textbf{DR}: how regular are your cycles usually?

\textbf{PT}: they can sometimes be off by a couple day’s, give or take because i have hypothyroidism and am taking synthroid. but as of lately with my last two cycles, they had predicted to the day or a day late.
 
\textbf{DR}: okay . Is this the first time you are missing period this late?

\textbf{PT}: no . I’ve had it be late by two weeks and even have missed it twice.

\textbf{DR}: okay. Have you been trying to lose weight?

\textbf{PT}: I’ve been watching what I’ve been eating, so yes .
 
\textbf{DR}: any recent change in your physical activity?

\textbf{PT}: no

\textbf{DR}: when was the last time you had your thyroid panel checked/tested ?

\textbf{PT}: just last week. everything is as normal as can be.

\textbf{DR}: that's great to know.

\\
\hline

\textbf{Model Output Summary}

\begin{itemize}
\item period date feb 9th to 13th. average cycle of about 32-33 days

\item no hormonal form of birth control apart from plan b.

\item they can sometimes be off by a couple day. has hypothyroidism and am taking synthroid. has predicted to the day or a day late.

\item had it be late by two weeks and even have missed twice.

\item trying to lose weight. been watching what is eating, so yes.

\item no recent change in physical activity

\item had thyroid panel checked/tested just last week. everything is as normal.

\end{itemize} \\
\hline

\textbf{Reference}

\begin{itemize}
\item last menstrual period is february 9th-13th. has average cycles of 32-33 days.

\item not on hormonal form of birth control apart from plan b.

\item cycles may be off by a couple days because of hypothyroidism and is taking synthyroid. since last 2 cycles they are predicted to the day or a day late.

\item not the first time missing period. has it late by 2 weeks and even missed it twice.

\item is watching what he or she eats to loose weight.

\item no recent chnages in physical activity.

\item checked thyroid panel last week and everything is normal.

\end{itemize} \\
\hline

\end{tabular}
\caption{Example Medical Dialogue and Output Model Summary.}
\label{fig:supp2}
\end{figure*}

\begin{figure*}[!]
\renewcommand{\arraystretch}{1.5}
\begin{tabular}{ |m{\linewidth}| } 
\hline
\textbf{DR}: For a while as in few weeks or months?

\textbf{PT}: I noticed it maybe a month ago I don't look at my chest that often because I've got a lot of dysphoria so I'm not really sure how long it 's been there

\textbf{DR}: and is it increasing in size?

\textbf{PT}: it's pretty big already so I can't tell

\textbf{DR}: Thank you. Are you able to hold the lump under skin with your two fingers?

\textbf{PT}: not really it seems attached to my sternum or whatever is under there

\textbf{DR}: How long have you been lifting heavy weights?

\textbf{PT}: like a month, I took a break this week though
 
\textbf{DR}: thanks for letting me know. Is this the first time you are lifting heavy weights?

\textbf{PT}: I haven't excersized since high school so yeah it's been like 10 years

\textbf{DR}: Alright. Does anything make the lump smaller or bigger? Say, for example, bending down, taking deep breaths, pressing on it, etc? 

\textbf{PT}: the visibility of the lump changes when I move around but not necessarily the size in general

\\
\hline

\textbf{Model Output Summary}

\begin{itemize}

\item noticed it maybe a month ago . don't look at chest that often because 've got a lot of dysphoria. not sure how long it's been there in months .

\item it's pretty big in size. can't tell it 's so increasing in size. 

\item not able to hold the lump under skin with two fingers. seems attached to sternum, whatever is under there

\item lifting heavy weights a month , took break this week

\item haven't excersized since high school and it 's been like 10 years.

\item lump changes when move around but not necessarily the size in general.

\end{itemize} \\
\hline

\textbf{Reference}

\begin{itemize}
\item noticed it a month ago . does not look at chest often because has dysphoria. not sure how long it has been there.

\item it is pretty big and can not tell if it is increasing in size.

\item not able to hold the lump under skin with two fingers. it seem attached to sternum.

\item is lifing heavy weights since a month . took a break this week.

\item has not exercised since high school. lifted heavy weights after 10 years.

\item no change in lump size in general but visibility changes with moving around.

\end{itemize} \\
\hline

\end{tabular}
\caption{Example Medical Dialogue and Output Model Summary.}
\label{fig:supp3}
\end{figure*}

\begin{figure*}[!]
\renewcommand{\arraystretch}{1.5}
\begin{tabular}{ |m{\linewidth}| } 
\hline
\textbf{DR}: Hi this is Dr. Cole and I am board certified in internal medicine. I am happy to help address your health concerns today. I have reviewed your chart and am wondering if you have anything to add or any specific questions for me?

\textbf{PT}: not for now. just the extreme fatigue and general feeling of being “off” nice to meet you.

\textbf{DR}: I am sorry you are experiencing this. Why are you taking propranolol?

\textbf{PT}: it was for migraines but I stopped taking it on tuesday I'm sorry, stopped taking it tuesday*
\\
\hline

\textbf{Model Output Summary}

\begin{itemize}
\item not for now. just the extreme fatigue and general feeling off".

\item taking propranolol for migraines but stopped taking it on tuesday . stopped taking it tuesday*

\end{itemize} \\
\hline

\textbf{Reference}

\begin{itemize}
\item has extreme fatigue and general feeling of being off.

\item taking propranolol for migraines and stopped taking it tuesday.

\end{itemize} \\
\hline

\end{tabular}
\caption{Example Medical Dialogue and Output Model Summary.}
\label{fig:supp1}
\end{figure*}

\begin{table*}[]

\centering
\resizebox{\textwidth}{!}{\begin{tabular}{lll}
%\toprule
\hline

\textbf{Snippet}   & \textbf{\baseline} & \textbf{\threem} \\ 
\hline

\begin{tabular}[c]{@{}l@{}}
DOCTOR: do you have chest pain while coughing? \\
PATIENT: yes and breathing.  \\
  \hspace{.7in}pretty much all the time \\
\end{tabular}  &
\begin{tabular}[c]{@{}l@{}}
\textbf{\textcolor{red}{no chest pain}} while coughing. \\pretty much all time.
\end{tabular}  &
\begin{tabular}[c]{@{}l@{}}
\textbf{\textcolor{blue}{has chest pain}} while coughing \\ a and breathing. 
\end{tabular}
\\ \hline

\begin{tabular}[c]{@{}l@{}}
DOCTOR: did you take any medication for it? \\
PATIENT: no 
\end{tabular}  &
\begin{tabular}[c]{@{}l@{}}
\textbf{\textcolor{red}{took medication}} for it.
\end{tabular}  &
\begin{tabular}[c]{@{}l@{}}
\textbf{\textcolor{blue}{[NO] not taking any medication}} for it

\end{tabular}
\\ \hline

\end{tabular}}
\caption{Improvements on Negatives and Affirmatives with \threem compared to \baseline}
\label{tab:qual3m}
\end{table*}

\end{document}